%% file: main.tex
\documentclass{article}    
\usepackage[final]{corl_2019} 
\usepackage[utf8]{inputenc}

\usepackage{epsfig} %
\usepackage{sidecap}

\usepackage{floatrow}

\usepackage{mathptmx} 
\usepackage{times} 
\usepackage{amsmath} 
\usepackage{amssymb}  
\usepackage{mathrsfs}
\usepackage{caption}  
\usepackage{subcaption}
\usepackage[numbers]{natbib}
\usepackage{algpseudocode}
\usepackage{algorithm}
\usepackage{bbm}
\usepackage{tikz}
\usetikzlibrary{shapes, arrows, automata, plotmarks}

\DeclareMathAlphabet{\mathcal}{OMS}{cmsy}{m}{n}

\input{mysymbol.sty}

\def\Tr{\mathsf{T}}

\renewcommand \red[1]         {}

\begin{document}

\title{Learning Decentralized Controllers for \\ Robot Swarms 
with Graph Neural Networks}

\author{
  Ekaterina Tolstaya\\
  Department of Electrical and Systems Eng.\\
  University of Pennsylvania,  USA\\
  \texttt{eig@seas.upenn.edu} \\
   \And
Fernando Gama \\
     Department of Electrical and Systems Eng.\\
  University of Pennsylvania, USA \\
  \texttt{fgama@seas.upenn.edu} \\
   \And 
   James Paulos \\
     Department of Mechanical Eng.\\
  University of Pennsylvania, USA \\
    \texttt{jpaulos@seas.upenn.edu} \\
   \And 
   George Pappas \\
     Department of Electrical and Systems Eng.\\
  University of Pennsylvania, USA \\
    \texttt{pappasg@seas.upenn.edu} \\
   \And 
   Vijay Kumar \\
     Department of Mechanical Eng.\\
  University of Pennsylvania, USA \\
      \texttt{kumar@seas.upenn.edu} \\
   \And
   Alejandro Ribeiro \\
     Department of Electrical and Systems Eng.\\
  University of Pennsylvania, USA \\
    \texttt{aribeiro@seas.upenn.edu} \\
}


\maketitle
\begin{abstract}
We consider the problem of finding distributed controllers for large networks of mobile robots with interacting dynamics and sparsely available communications.
Our approach is to learn local controllers that require only local information and communications at test time by imitating the policy of centralized controllers using global information at training time. 
By extending aggregation graph neural networks to time varying signals and time varying network support, we learn a single common local controller which exploits information from distant teammates using only local communication interchanges.
We apply this approach to the problem of flocking to demonstrate performance on communication graphs that change as the robots move. We examine how a decreasing communication radius and faster velocities increase the value of multi-hop information.
\end{abstract}


\input{intro.tex}
\input{system.tex}

\input{agg_gnn.tex}

\input{flocking.tex}

\input{conclusion.tex}


\acknowledgments{Supported by the Semiconductor Research Corporation (SRC) and DARPA, ARL DCIST CRA W911NF-17-2-0181, NSF DGE-1321851, NSF CNS-1521617,  and the Intel Science and Technology Center for Wireless Autonomous Systems (ISTC-WAS).}
\bibliography{myIEEEabrv,main}

\newpage
\clearpage
\input{appendix.tex}

\end{document}

%% file: intro.tex
\section{Introduction}

Large scale swarms are composed of multiple agents that collaborate to accomplish a task. In the future, these systems could be deployed to, for example, provide on-demand wireless networks \cite{sharma2016uav}, perform rapid environmental mapping \cite{thrun2000real, thrun2005multi}, search after natural disasters \cite{baxter2007multi, jennings1997cooperative}, or enable sensor coverage in cluttered and communications denied environments \cite{zhang2005maintaining}. For as long as scale is moderate, the group can be controlled as a whole from a central agent. However, as we reach for larger number of agents, decentralized control becomes a necessity. Individual agents must decide on control actions that are conducive to accomplishing a collective task from their local observations and communication with nearby peers. It has long been known that finding optimal controllers in these distributed settings is challenging \cite{witsenhausen1968counterexample}. This motivates the use of heuristics in general and, in particular and as we advocate in this paper, the use of learned heuristics.

We emphasize that the challenge in decentralized control stems from the local information structure generated by the unavoidable restriction to communicate with only nearby agents. 
Building on this observation, we propose the use of imitation learning to train policies that respect the local information structure of a distributed system, while attempting to mimic the global policy of a clairvoyant expert (Section \ref{sec_control_of_networked_systems}). The value of imitation learning has been demonstrated in single agent robotics problems such as autonomous driving \cite{pomerleau1989alvinn} and quadrotor navigation \cite{giusti2016machine}. When designing multi-agent systems, we must contend with the dimensionality growth of the system as new agents are added. 

Both of these problems can be overcome if we use a Graph Neural Networks \cite{bruna14-deepspectralnetworks, defferrard17-cnngraphs, kipf17-classifgcnn, gama2018convolutional, ruiz19nonlinear}. In particular, aggregation graph neural networks (aggregation GNNs) \cite{gama2018convolutional} are especially suited to teams of agents operating over a physical network because the architecture operates in an entirely local fashion involving only communication between nearby neighbors. In small-size multi-agent problems \cite{tacchetti2019relational}, it is shown that explicitly learning the graph of agent relationships aids in distilling decentralized agent policies from expert policies trained using Actor-Critic methods. In contrast, we leverage known relationships and connectivity between agents in order to extract features with graph convolutions, following the approach of \cite{gama2018convolutional}. Exploiting the known network structure allows us to consider teams an order of magnitude larger and highlights the value of using information from multi-hop neighbors.

Related work in multiagent learning for control in imitation \cite{paulos2019decentralization} and reinforcement \cite{huttenrauch2019deep} settings address varying neighbor relationships in communication but provide no mechanism for multi-hop information flow.
Explicit multi-hop message passing between all team members allows offline \cite{caro2013simple} and incremental \cite{giusti2012cooperative} training for group inference but incurs a superlinear growth in communications with team size.
Scalable online training is achieved in \cite{otte2018emergent} by viewing the team as a distributed neural network, but training and inference must cease once agents move and the network changes.

We examine flocking tasks to highlight the ability of our approach to handle dynamic communication networks.
Flocking has natural extensions to transportation and platooning of autonomous vehicles where agents rely on local observations to align their velocities and regulate their spacing \cite{tanner2004flocking}.
Previous work focuses on developing local controllers which incorporate only observations from immediate neighbors \cite{reynolds1987flocks,jadbabaie2003coordination,tanner2003stable}.
We show that a global controller inspired by \cite{tanner2003stable} outperforms such local controllers, but global approaches are not practical for real deployments.
The novelty of our approach to flocking is to aggregate from multi-hop neighbors; this ability allows us to approach the performance of global solutions while respecting realistic communication constraints.
Prior to this work, there has been no principled approach for augmenting the communication between neighbors to pass on information aggregated from multi-hop neighbors.

{\bf Notation.} For a matrix $\bbA$ with entries $a_{ij}$ we use $[\bbA]_{ij} = a_{ij}$ to represent its $(i,j)$th entry and $[\bbA]_{i} = [a_{i1},\ldots,a_{iN}]$ to represent its $i$th row. For matrices $\bbX$ and $\bbY$ we use $[\bbX,\bbY]$ to represent its block column concatenation and $[\bbX;\bbY]$ for its block row stacking. 

%% file: system.tex
\section{Control of Networked Systems}\label{sec_control_of_networked_systems}

We consider a team of $N$ agents distributed in space and tasked with controlling some large scale dynamical process.
We characterize this dynamical process by the collection of state values $\bbx_{i}(t)\in\reals^p$ observed at the locations of each agent $i$ at time $t$, as well as the control actions $\bbu_{i}(t)\in\reals^q$ that agents take.
Grouping local states into the joint system state matrix $\bbx(t) = [\bbx_{1}^{\Tr}(t);\ldots;\bbx_{N}^{\Tr}(t)]\in\reals^{N\times p}$, and local actions into the overall system action $\bbu(t) = [\bbu_{1}^{\Tr}(t);\ldots;\bbu_{N}^{\Tr}(t)]\in\reals^{N\times q}$, we can write the evolution of the dynamical process through a differential equation of the form, $ \dot{\bbx}(t) = f\left(\bbx(t),\bbu(t)\right)$.


In order to design a controller for this dynamical system we operate in discrete time by introducing a sampling time $T_s$ and a discrete time index $n$.
We define $\bbx_{n} = \bbx(nT_s)$ as the discrete time state and $\bbu_{n}$ as the control action held from time $nT_s$ until time $(n+1)T_s$.
Solving the differential equation between times $nT_s$ and $(n+1)T_s$, we end up with the discrete dynamical system 
\begin{equation} \label{eq:discreteDynamic}
  \bbx_{n+1} \;= \;\int_{nT_s}^{(n+1)T_s} f\left(\bbx(t),\bbu_{n}\right)\,\ dt + \bbx_n, \;\;\;
  \text{with\ }  \bbx(nT_s) = \bbx_{n} .
\end{equation}
At each point in (discrete) time, we consider a cost function $c(\bbx_n, \bbu_n)$.
The objective of the control system is to choose actions $\bbu_n$ that reduce the accumulated cost $\sum_{n=0}^\infty c(\bbx_n, \bbu_n)$.
When the collection of state observations $\bbx_{n} = [\bbx_{1n}^{\Tr};\ldots;\bbx_{Nn}^{\Tr}]$ is available at a central location it is possible for us to consider centralized policies $\pi$ that choose control actions $\bbu_n = \pi(\bbx_n)$ that depend on global information.
In such case the optimal policy is the one that minimizes the expected long term cost.
%
%
\red{The global flocking controller isn't optimal wrt. the cost function}
If the dynamics in $f(\bbx(t),\bbu(t))$  and the costs $c \big(\bbx_n, \pi(\bbx_n) \big)$ are known, as we assume here, there are several techniques to find the optimal policy $\pi^*$ \cite{zhou1996robust, bemporad2002explicit}.
In this paper we are interested in decentralized controllers that operate without access to global information and interpret the optimal controller as a benchmark that decentralized controllers are trying to imitate. 


The agent network is described by the presence of an edge $(i,j)\in\ccalE_n$ which indicates that $j$ may send data to $i$ at time $n$.
When this happens we say that $j$ is a neighbor of $i$ and define the neighborhood of $i$ at time $n$ as the collection of all its neighbors, $\ccalN_{in} = \{ j: (i,j)\in\ccalE_n \}$.
%
%
We can also define multi-hop neighborhoods of a node.
We first define the 0-hop neighborhood of $i$ to be the node itself, $\ccalN_{in}^{0} = \{i\}$.
The 1-hop neighborhood of $i$ is defined as simply the neighborhood of $i$, $\ccalN_{in}^{1} = \ccalN_{in}$
We can now define the $k$-hop neighborhood of $i$ as the set of nodes that can reach node $i$ in exactly $k$ hops. 
%
%
The $k$-hop neighbors of $i$ are the nodes that are $(k-1)$-hop neighbors at time $n-1$ of the neighbors of $i$ at time $n$. This recursive neighborhood definition characterizes the information that is available to node $i$ at time $n$. This information includes the local state $\bbx_{in}$ that can be directly observed by node $i$ at time $n$ as well as the value of the state $\bbx_{j(n-1)}$ at time $(n-1)$ for all nodes $j$ that are 1-hop neighbors of $i$ at time $n$ since this information can be communicated to node $i$. Node $i$ can also learn the state $\bbx_{j(n-2)}$ of 2-hop neighbors at time $n-2$ since that information can be relayed from neighbors. In general, Eqn. \ref{eq:eqn_history_available_at_node_n} defines the information history $\ccalH_{in}$ of node $i$ at time $n$ as the collection of state observations out to a maximal history depth $K$.
\begin{equation}
\label{eq:eqn_history_available_at_node_n}
   \ccalN_{in}^{k} = \Big\{ j' \in \ccalN_{j (n-1)}^{k-1} : j \in \ccalN_{in} \Big\}, \;\;\;\;\;\;
  \ccalH_{in} = \bigcup_{k=0}^{K-1} 
      \Big\{ \bbx_{j (n-k)} : j \in  \ccalN_{in}^{k} \Big \}
\end{equation}
The decentralized control problem consists on finding a policy that minimizes the long term cost $\sum_{n=0}^\infty c(\bbx_n, \bbu_n)$ restricted to the information structure in \eqref{eq:eqn_history_available_at_node_n}. This leads to problems in which finding optimal controllers is famously difficult to solve \cite{witsenhausen1968counterexample} except in some particularly simple scenarios \cite{eksin2013bayesian}. This complexity motivates the introduction of a method that learns to mimic the global controller. 
Formally, we introduce a parameterized policy $\pi(\ccalH_{in},\bbH)$ that maps local information histories $\ccalH_{in}$ to local actions $\bbu_{in}=\pi(\ccalH_{in},\bbH)$ as well as a loss function  $\ccalL(\pi, \pi^*)$ to measure the difference between  the optimal centralized policy $\pi^*$ and a system where all agents (locally) execute the (local) policy $\pi(\ccalH_{in},\bbH)$. Our goal is to find the tensor of parameter $\bbH$ that solves the optimization problem 
\begin{equation} \label{eq:decentralizedController}
    \bbH^* = \argmin_{\bbH} 
                  \mathbb{E}^{\pi^*} \Big[  \ccalL\Big( \pi\big(\ccalH_{in},\bbH\big), \pi^*(\bbx_n) \Big) \Big],
\end{equation}
where we use the notation $\mathbb{E}^{\pi^*}$ to emphasize that the distribution of observed states $\bbx_n$ over which we compare the policies $\pi\big(\ccalH_{in},\bbH\big)$ and $\pi^*(\bbx_n)$ is that of a system that follows the optimal policy $\pi^*$. The formulation in \eqref{eq:decentralizedController} is one in which we want to learn a policy $\pi\big(\ccalH_{in},\bbH\big)$ that mimics $\pi^*$ to the extent that this is possible with the information that is available to each individual node. The success of this effort depends on the appropriate choice of the parameterization that determines the family of policies $\pi(\ccalH_{in},\bbH)$ that can be represented by different choices of parameters, $\bbH$. In this work, we advocate the use of an aggregation graph neural network (Section \ref{sec_agg_gnn}) and demonstrate its applications to the problem of flocking with collision avoidance (Section \ref{sec_flocking}).

%% file: agg_gnn.tex
\section{Delayed Aggregation Graph Neural Networks}\label{sec_agg_gnn}

Aggregation graph neural networks (aggregation GNNs) are information processing architectures that operate on network data in a completely local and decentralized way, harnessing all the useful information by repeated exchanges with their neighbors \cite{gama2018convolutional}. This makes them suitable choices for parameterizing the decentralized policy $\pi(\ccalH_{in},\bbH)$ in \eqref{eq:decentralizedController}. However, aggregation GNNs operate on fixed graph signals defined over a fixed graph support, so in what follows, we extend aggregation GNNs to operate on time-varying graph processes defined over a time-varying graph support.

The graph support $\ccalG_{n}$ is conveniently described by a \emph{graph shift operator} $\bbS_{n} \in \reals^{N \times N}$ which is a matrix whose element $[\bbS_{n}]_{ij}$ can be nonzero only if $(j,i) \in \ccalE_{n}$ or if $i=j$, respecting the sparsity of the graph \cite{sandryhaila14-mag, chen15-selection, marques16-aggregation}.
It follows that \(\bbS_{n}\) defines linear, local operations since multiplications with $\bbS_{n}$ can be computed only with neighboring exchanges.
More precisely, recall that $[\bbx_{(n-1)}]_j = \bbx_{j(n-1)}^\Tr$ denotes the state observed by node $j$ at time $n-1$ and further distribute the product $\bbS_{n} \bbx_{n-1}$ so that $[\bbS_{n}\bbx_{n-1}]_{i}$ is associated with node $i$.
Since $[\bbS_{n}]_{ij}\neq0$ only if $(j,i)\in\ccalE_n$ or if $i=j$ we can write
\begin{equation} \label{eq:localCommunication}
    \Big[\,\bbS_{n} \bbx_{n-1}\,\Big]_{i} \ = 
           \sum_{j=i, j \in \ccalN_{in}} 
                  \Big[\,\bbS_{n}\,\Big]_{ij}\Big [\,\bbx_{n-1}\,\Big]_{j} \ .
\end{equation}
Thus, node $i$ can carry its part of the multiplication operation by receiving information from neighboring nodes.
Examples of valid shift operators include weighted and unweighted adjacency matrices as well as weighted, unweighted, or normalized Laplacians.
Aggregation GNNs leverage the locality of \eqref{eq:localCommunication} to build a sequence of recursive $k$-hop neighborhood [Eq. \eqref{eq:eqn_history_available_at_node_n}] aggregations to which a neural network can be applied. More precisely, consider a sequence of signals $\bby_{kn}\in\reals^{N\times p}$ that we define through the recursion
\begin{equation} \label{eq:ydef}
    \bby_{kn} = \bbS_n \bby_{(k-1)(n-1)} ,
\end{equation}
with the initialization $\bby_{0n} = \bbx_{n}$. If we fix the time $n$ and consider increasing values of $k$, the recursion in \eqref{eq:ydef} produces a sequence of signals where the first element is $\bby_{0n} = \bbx_{n}$, the second element is $\bby_{1n} = \bbS_n\bby_{0(n-1)} = \bbS_n\bbx_{n-1}$, and, in general, the $k$th element is $\bby_{kn} = (\bbS_n\bbS_{n-1}\ldots\bbS_{n-k+1})\bbx_{n-k}$. Thus, \eqref{eq:ydef} is modeling the diffusion of the state $\bbx_{n-k}$ through the sequence of time varying networks $\bbS_{n-k+1}$ through $\bbS_n$.
This diffusion can be alternatively interpreted as an aggregation; at node $i$ we can define an aggregation sequence of $K$ elements as
\begin{equation} \label{eq:z_def}
    \bbz_{in} = \Big[ \big[\bby_{0n}\big]_i \,;
                      \big[\bby_{1n}\big]_i \,;
                      \ldots\,; 
                      \big[\bby_{(K-1)n}\big]_i \Big].
\end{equation}
The first element of this sequence is $[\bby_{0n}]_i = [\bbx_{n}]_i = \bbx_{in}^\Tr$ which represents the local state of node $i$. The second element of this sequence is  $[\bby_{1n}]_i = [\bbS_n\bbx_{n-1}]_i$ which aggregates the states $\bbx_{j(n-1)}$ of 1-hop neighboring nodes $j\in\ccalN_{in}^{1}$ observed at time $n-1$ with a weighted average. In fact, this element is precisely the outcome of the local average shown in \eqref{eq:localCommunication}. If we now focus on the third element we see that $[\bby_{2n}]_i = [\bbS_n\bbS_{n-1}\bbx_{n-2}]_i$ is an average of the states $\bbx_{j(n-2)}$ of 2-hop neighbors $j\in\ccalN_{in}^{2}$ at time $n-2$. In general, the $k+1$st element of $\bbz_{in}$ is $[\bby_{kn}]_i = [\bbS_n\bbS_{n-1}\ldots\bbS_{n-k+1}\bbx_{n-k}]_i$ which is an average of the states $\bbx_{j(n-k)}$ of k-hop neighbors $j\in\ccalN_{in}^{k}$ observed at time $n-k$. From this explanation we conclude that the sequence $\bbz_{in}$ is constructed with state information that is part of the local history $\ccalH_{in}$ defined in \eqref{eq:eqn_history_available_at_node_n} and therefore a valid basis for a decentralized controller. We highlight in equations \eqref{eq:localCommunication} and  \eqref{eq:ydef} that agents forward the aggregation of their neighbors' information, rather than a list of neighbors' states, further along to multi-hop neighbors. 

An important property of the aggregation sequence $\bbz_{in}$ is that it exhibits a regular temporal structure as it is made up of nested aggregation neighborhoods. This regular structure allows for the application of a regular convolutional neural network (CNN) \cite{gama2018convolutional} of depth $L$, where for each layer $\ell=1,\ldots,L$, we have
\begin{equation} \label{eq:AggGNN}
\bbz^{(\ell)}_{in} = \sigma^{(\ell)} ( \bbH^{(\ell)} \bbz^{(\ell-1)}_{in} )
\end{equation}
with $\sigma^{(\ell)}$ a pointwise nonlinearity and $\bbH^{(\ell)}$ a bank of small-support filters containing the \emph{learnable} parameters. For each node $i$, we set $\bbz^{(0)}_{in} = \bbz_{in}$ and collect the output $\bbz^{(L)}_{in}=\bbu_{in}$ to be the decentralized control action at node $i$, at time $n$. We note that the filters $\bbH^{(\ell)}$ are shared across all nodes. We refer the reader to Section \ref{sec_gnn_details} in the supplementary materials for further details on aggregation GNNs.

The aggregation GNN architecture, described in equations \eqref{eq:ydef}-\eqref{eq:AggGNN}, constitutes a local parameterization of the policy $\pi(\ccalH_{in},\bbH)$ that exploits the network structure and involves communication exchanges only with neighboring nodes. To \emph{learn} the parameters for $\bbH$, we use a training set $\ccalT$ consisting of sample trajectories $(\bbx_{n},\pi^{\ast}(\bbx_{n}))$ obtained from the global controller $\bbu^{\ast}_{n} = \pi^{\ast}(\bbx_{n})$. 
We thus minimize the loss function over this training set, Eq. \eqref{eq:decentralizedController}, where $\bbu_{n}$ collects the output $\bbu_{in}=\bbz^{(L)}_{in}$ of \eqref{eq:AggGNN} at each node $i$:
\begin{equation} \label{eq:empirical_loss}
    \bbH^{\ast} = \argmin_{\bbH} \sum_{(\bbx_{n},\pi^{\ast}(\bbx_{n})) \in \ccalT}
                   \ccalL\Big( \bbu_{n}, \bbu_{n}^{\ast} \Big)
\end{equation}
%
We note that the policy learned from \eqref{eq:empirical_loss} can be extended to any network since the filters $\bbH^{(\ell)}$ can be applied independently of $\bbS_{n}$, facilitating transfer learning. This transfer is enabled by sharing the filter weights $\bbH$ among nodes at training time. The learned aggregation GNN models are, therefore, network and node independent. Graph covariance is an advantage for the applications examined in this work, but may not be a suitable assumption in all problems \cite{gama19-stability}.


%% file: flocking.tex
%
\section{Methods: Learning to Flock} \label{sec_flocking}

We evaluate the proposed methodology by learning a local controller for flocking with collision avoidance \cite{tanner2003stable}. Consider then a team of $N$ agents in which $\bbr_i$ denotes the position of agent $i$ and $\bbv_i = \dot{\bbr}_i$ denotes its velocity. We assume that accelerations are fully controllable and therefore attempt to design control inputs $\bbu_i$ so that the change in velocity of agent $i$ is $\dot{\bbv}_i = \bbu_i$. Further denote as $\bbr_{ij} := \bbr_j - \bbr_i$ the relative position of agent $j$ with respect to agent $i$ and introduce a constant $\rho>1$ to define the collision avoidance potential 
\begin{equation} \label{eq:potentialij}
U (\bbr_i, \bbr_j)
   \ =\ 1/\left\|\bbr_{ij}\right\|^2 
           \  + \  \log \left\|\bbr_{ij}\right\|^2
           \text{\ if\ } \left\|\bbr_{ij}\right\|<\rho;
           \quad
           1/\rho^2 + \log(\rho^2) \text{\ otherwise.}
\end{equation}
Since in addition to avoid collisions we want all agents to coordinate on their velocities we propose the controller
\begin{equation} \label{eq:controller_global}
   \bbu_i^* = - \sum_{j =1  }^N (\bbv_i - \bbv_j) - \sum_{j =1}^N \nabla_{\bbr_i} U(\bbr_i, \bbr_j).
\end{equation} 
The combination of the collision avoidance potential with the velocity agreement term  $\bbv_i - \bbv_j$ pushes the agents to coordinate their velocities while avoiding collisions. Observe that the potential $U (\bbr_i, \bbr_j)$ diverges at $\|r_{ij}\| = 0$,
has a minimum when the distance between agents is $\|r_{ij}\| = 1$ and is indifferent when $\|r_{ij}\| >\rho$. It therefore pushes agents to either be at distance $\|r_{ij}\| = 1$ of each other or at distance $\|r_{ij}\| > \rho$ of each other. 

The controller in \eqref{eq:controller_global} requires access to all agent velocities and all agent positions. In the parlance of Section \ref{sec_control_of_networked_systems}, it is a clairvoyant centralized controller. In practice, agents can have access to local information only as they can only sense and communicate with agents within distance $ \|\bbr_{ij} \| < R$ of each other. A controller that respects the locality of sensing and communication is
\begin{equation} \label{eq:controller_local}
   \bbu_i^\dagger = - \sum_{j \in \mathcal{N}_i} (\bbv_i - \bbv_j) - \sum_{j \in \mathcal{N}_i} \nabla_{\bbr_i} U(\bbr_i, \bbr_j).
\end{equation} 
The controller in \eqref{eq:controller_local} differs from the centralized controller in \eqref{eq:controller_global} in that the sums are over agents $j\in\ccalN_i$ defined as those whose distance to agent  $i$ is $\|\bbr_{ij} \| < R$ for some communication radius $R$. For both the centralized and decentralized controllers, we have chosen $R=\rho$. It follows that the controller in \eqref{eq:controller_local} is decentralized as it can be implemented by accessing local information only. The controllers in \eqref{eq:controller_global} and \eqref{eq:controller_local} have the same stationary points but \eqref{eq:controller_local} may -- indeed, it most often does -- take longer to coordinate agent velocities. In the next section we use an aggregation GNN to learn a local controller that mimics \eqref{eq:controller_global}. We will show that this learned controller outperforms \eqref{eq:controller_local} and achieves a performance that is similar to \eqref{eq:controller_global}.

%
%
%
%

%


The global and local controllers are both non-linear in the states of the agents. The classical aggregation GNN approach does not allow for non-linear operations prior to aggregation, so we cannot use the position and velocity vectors alone to imitate the global controller. Rather than directly using the state $[\bbr_i, \bbv_i]$ of each node $i$ for aggregation, we design the relevant features needed to replicate the non-linear controller using only a linear aggregation operation, where $[\bbx_n]_i \in \mathbb{R}^6$:
\begin{equation} \label{eq:flockfeatures}
[\bbx_n]_i = \Bigg\lbrack
\sum_{j \in \mathcal{N}_i} (\bbv_{i,n} - \bbv_{j,n}), \quad
\sum_{j \in \mathcal{N}_i} \frac{\bbr_{ij, n}}{\left\|\bbr_{ij,n}\right\|^4}, \quad
\sum_{j \in \mathcal{N}_i} \frac{\bbr_{ij,n}}{ \left\|\bbr_{ij,n} \right\|^2}
\Bigg\rbrack
\end{equation}
This observation vector is then used during aggregation as described in \eqref{eq:ydef}-\eqref{eq:z_def}.
The local controller also requires the computation of \eqref{eq:flockfeatures}, so we are not giving the GNN an unfair advantage by providing the instantaneous measurements of neighbors' states. We quantify the cost of a trajectory by the variance in velocities, where $T$ is the number of time steps in the trajectory. The variance in velocities measures how far the system is from consensus in velocities \cite{xiao2007distributed}: %
\begin{equation} \label{eq:flockcost}
C = \frac{1}{N} \sum_{n=1}^T  \sum_{j =1 }^N \left\| \bbv_{j,n} - \frac{1}{N} \bigg[  \sum_{i =1}^N \bbv_{i,n} \bigg]  \right\|^2.
\end{equation}

For our baseline scenario we consider a flock of $N = 100$ agents with a communication radius of $R= \rho = 1.0$ m and a discretization time period of $T_s=0.01$ s.
The flock locations were initialized uniformly on the disc with radius $\sqrt{N}$ to normalize the density of agents for changing flock sizes.
Initial agent velocities are controlled by a parameter $v_{init} = 3.0$ m/s: agent velocities are sampled uniformly from the interval $[-v_{init}, +v_{init}]$ and then a bias for the whole flock is added, also sampled from $[-v_{init}, +v_{init}]$.
To eliminate unsolvable cases, configurations are resampled if any agent fails to have at least two neighbors or if agents begin closer than 0.1 m.
Finally, acceleration commands are saturated to the range $[-100, 100]$ m/s$^2$ to improve the numerical stability of training.

To obtain an estimate of the action to take, each agent operates an aggregation GNN architecture, as described in Section III, with input features given by \eqref{eq:flockfeatures}. The aggregated vectors $\bbz_{in}$ in Eq. \eqref{eq:z_def} are built from $K-1$ neighbor exchanges, and then fed into a fully connected neural network as per Eq. \eqref{eq:AggGNN}, with two hidden layers of $32$ neurons each and a Tanh activation function. The network was implemented in PyTorch and trained over a MSE cost function, using the Adam optimizer with learning rate $5 \cdot 10^{-5}$ and forgetting factors $\beta_{1}=0.9$ and $\beta_{2}=0.999$. Each model was trained using $400$ trajectories, each of length $200$ steps total. For testing, $20$ trajectories of length $200$ were observed by using the learned controller only.

In practice, following the optimal policy to collect training data results in a distribution of states that is not representative of those seen at test time.
To resolve this we use the Dataset Aggregation (DAgger) algorithm and follow the learner's policy instead of the expert's with probability $1\!-\!\beta$ when collecting training trajectories \cite{ross2011reduction}.
The probability \(\beta\) of choosing the expert action while training is decayed by a factor of \(0.993\) after each trajectory to a minimum of \(0.5\).

\begin{figure}[t]
  \setlength{\abovecaptionskip}{-2pt}
  \setlength{\belowcaptionskip}{-1pt}
\centering
\begin{subfigure}[b]{.4\linewidth}
\abovecaptionskip -2pt
\includegraphics[width=\textwidth]{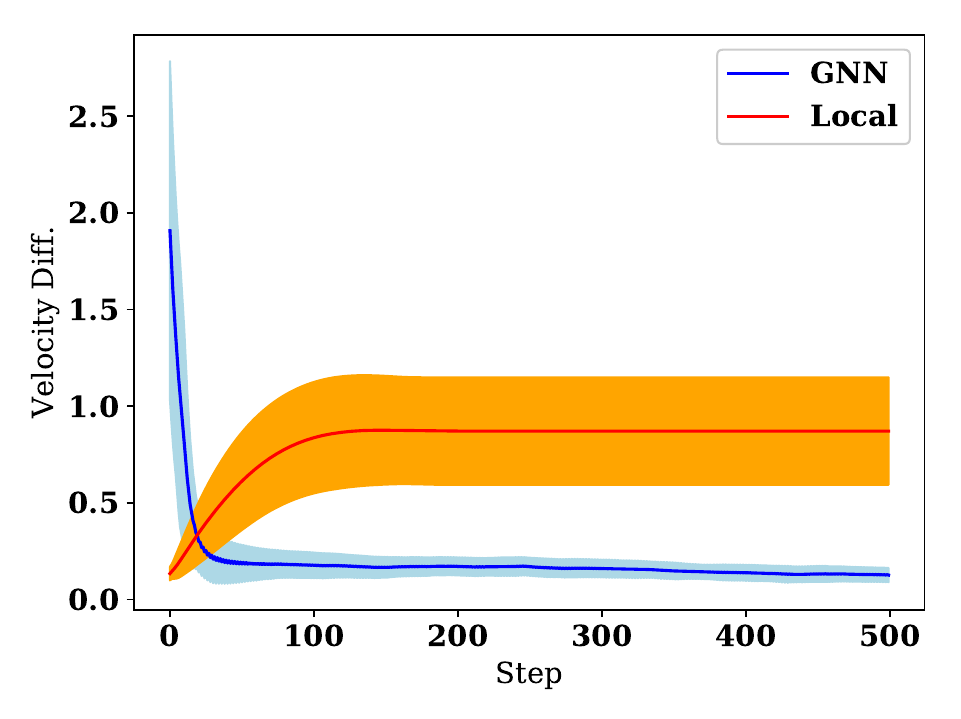}
\caption{Average difference in velocities}\label{fig:traj_vel}
\end{subfigure}
\begin{subfigure}[b]{.4\linewidth}
\abovecaptionskip -2pt
\includegraphics[width=\textwidth]{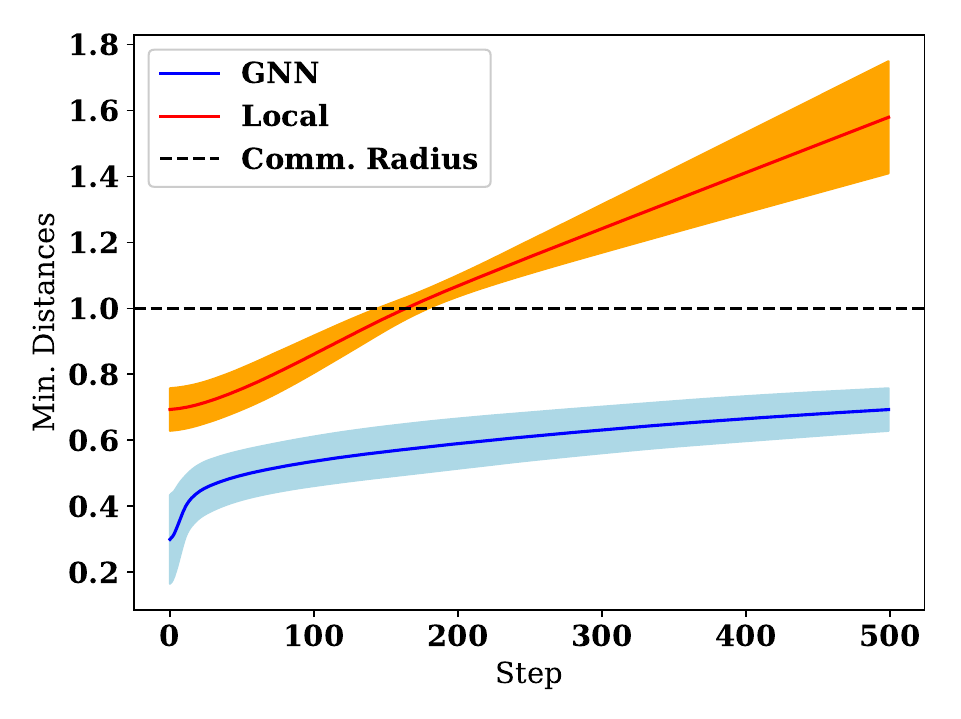}
\caption{Average minimum distance to a neighbor }\label{fig:traj_dist}
\end{subfigure}
\begin{subfigure}[b]{.4\linewidth}
\abovecaptionskip -2pt
\includegraphics[width=\textwidth]{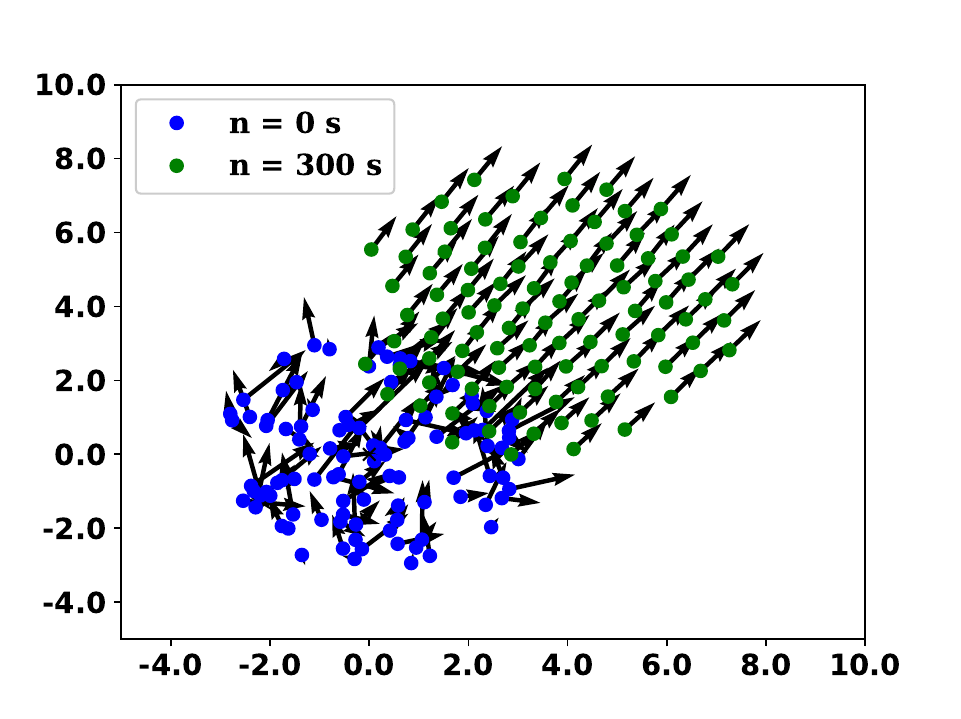}
\caption{Flock positions using the GNN}\label{fig:traj_300}
\end{subfigure}
\begin{subfigure}[b]{.4\linewidth}
\abovecaptionskip -2pt
\includegraphics[width=\textwidth]{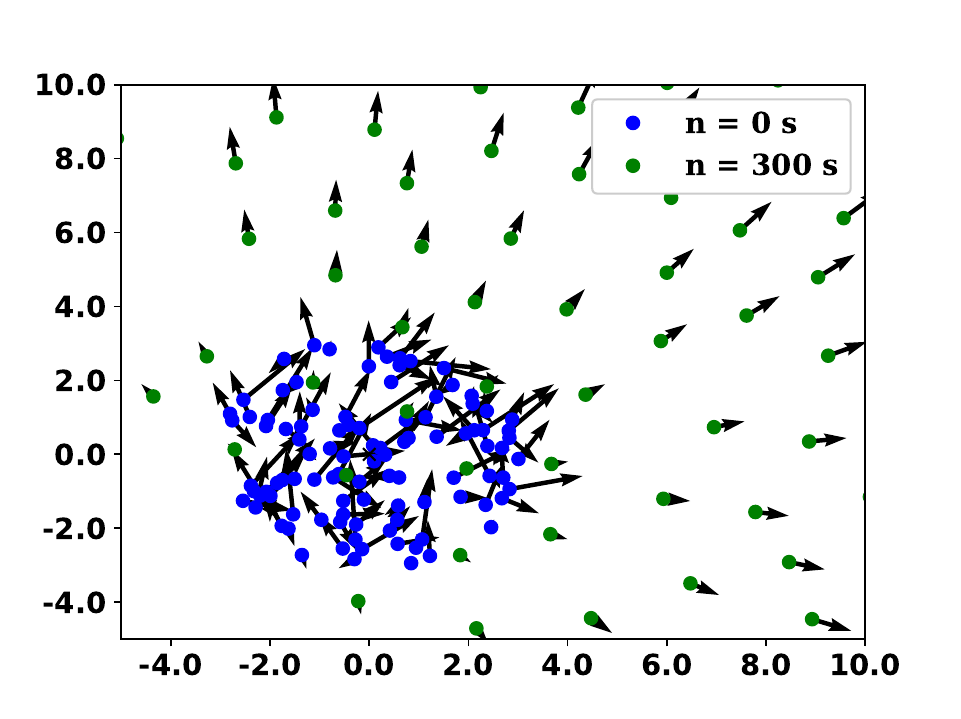}
\caption{Flock positions using the local controller}\label{fig:dec_traj_300}
\end{subfigure}
\caption{ 
The GNN ($K=3$) maintains a cohesive flock, while the local controller allows the flock to scatter.
\vspace{-0.5cm}
}
\label{fig:flocking_traj}
\end{figure}

\section{Results} \label{sec_results}

We report results comparing \eqref{eq:controller_local} and \eqref{eq:controller_global} for point masses with fully controllable accelerations in Section \ref{subsec:systemParams}. This simple setting allows for an exploration of the effect of different system parameters such as initial velocity or communication radius, to determine experimentally the scenarios on which the aggregation GNN offers good performance. In Section~\ref{subsec:transfer} we study the case of transfer learning, where we train the model in one network but test it in another (for example, with different number of agents), and also by exporting the trained architecture to other physical models beyond the point-mass model, as shown in the AirSim simulator (Sec.~\ref{subsec:airsim}).

\subsection{Learning to flock with point masses} \label{subsec:systemParams}


First, we compare the performance of the GNN controller with $K=3$ to the local controller $\bbu_{i}^{\dagger}$ \eqref{eq:controller_local}.
Figure \ref{fig:traj_vel} depicts the magnitude of velocity differences between agents over the course of a trajectory in terms of the population mean and standard devision.
The GNN converges much more rapidly and, unlike the local controller, approaches a perfect velocity consensus.
Part of the reason for this is explained in Figure \ref{fig:traj_dist}, which plots agents' minimum distance to any neighbor over time.
The GNN control approaches a uniform flock spacing, but the local controller fails to stop agents from dispersing quickly enough.
Soon the local controller's network has become completely disconnected as agent distances exceed their communication range of \(R=1\).
One flock trajectory is depicted for the GNN controller in Fig.~\ref{fig:traj_300} and the local controller in Fig.~\ref{fig:dec_traj_300}.
Each diagram shows the initial agent positions and velocities at time $n=0$ and then at $n=300$, qualitatively illustrating the stable flocking of the GNN and failure of the local controller.

\begin{figure}[t]
  \setlength{\abovecaptionskip}{-2pt}
\centering
\begin{subfigure}[t]{.4\linewidth}
\includegraphics[width=\textwidth]{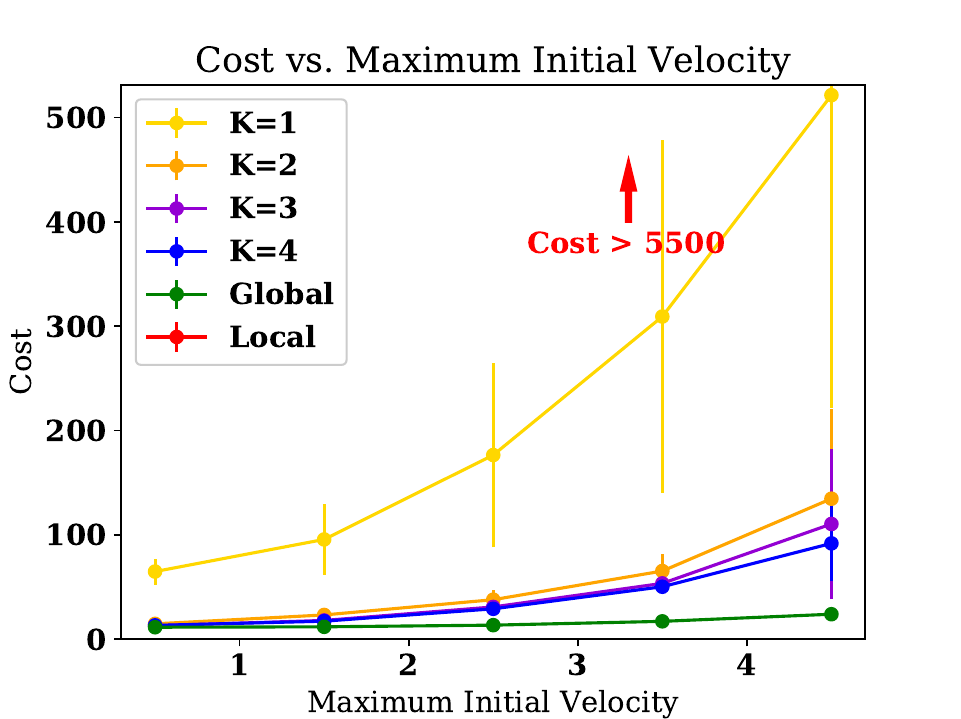} 
\subcaption{ } 
\label{fig:vel}
\end{subfigure}%
\begin{subfigure}[t]{.4\linewidth}
\includegraphics[width=\textwidth]{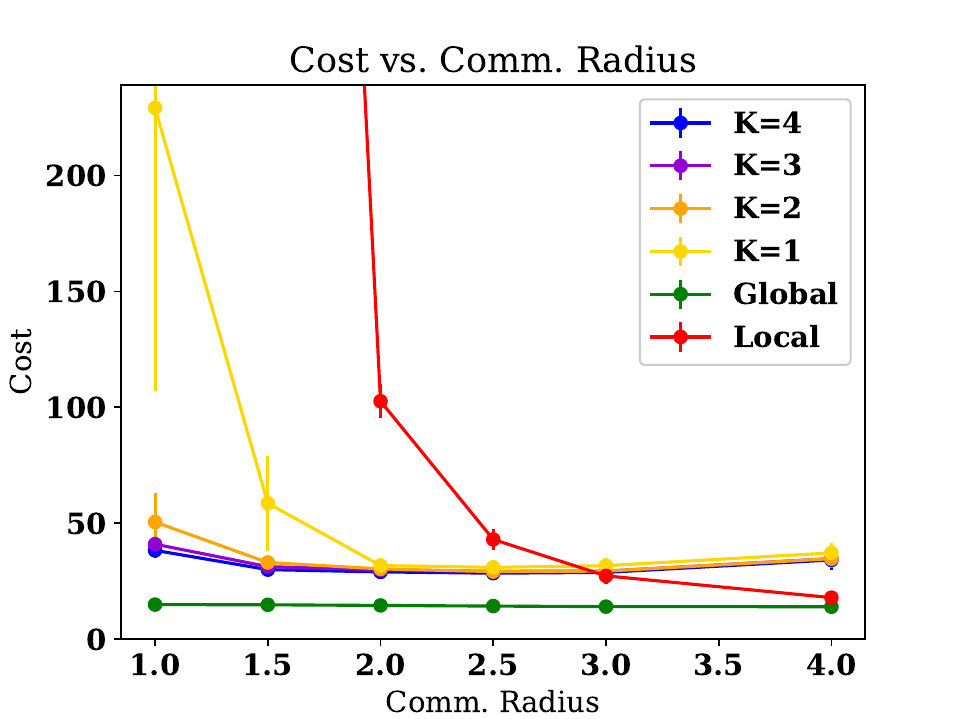}
\subcaption{ } 
\label{fig:rad}
\end{subfigure}
\begin{subfigure}[t]{.4\linewidth}
\includegraphics[width=\textwidth]{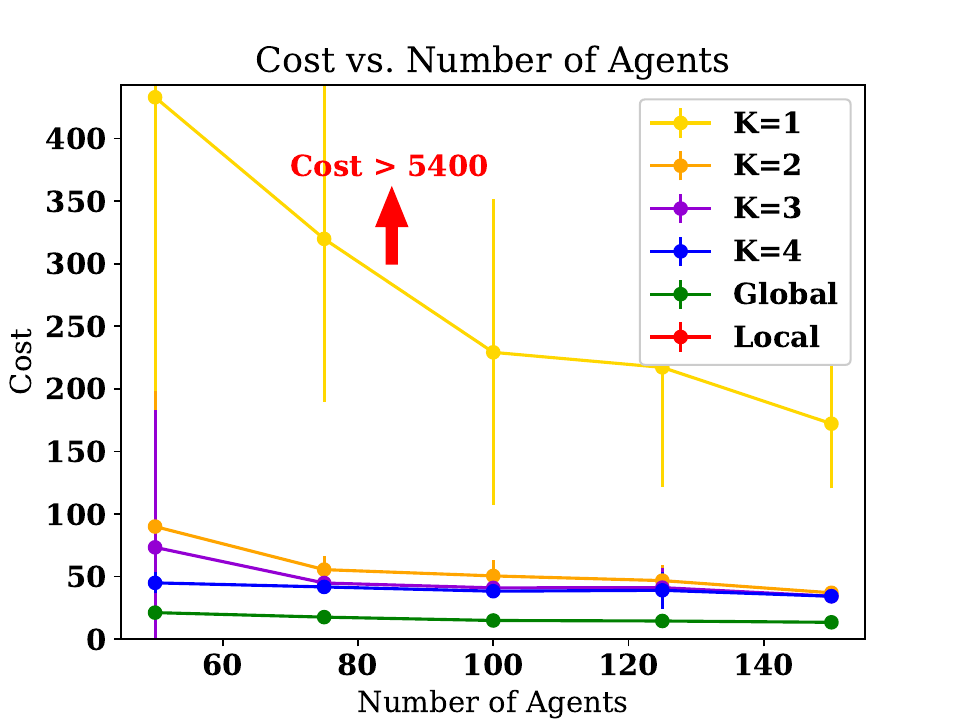} 
\subcaption{ }
\label{fig:flocking_n}
\end{subfigure}%
\begin{subfigure}[t]{.4\linewidth}
\includegraphics[width=\textwidth]{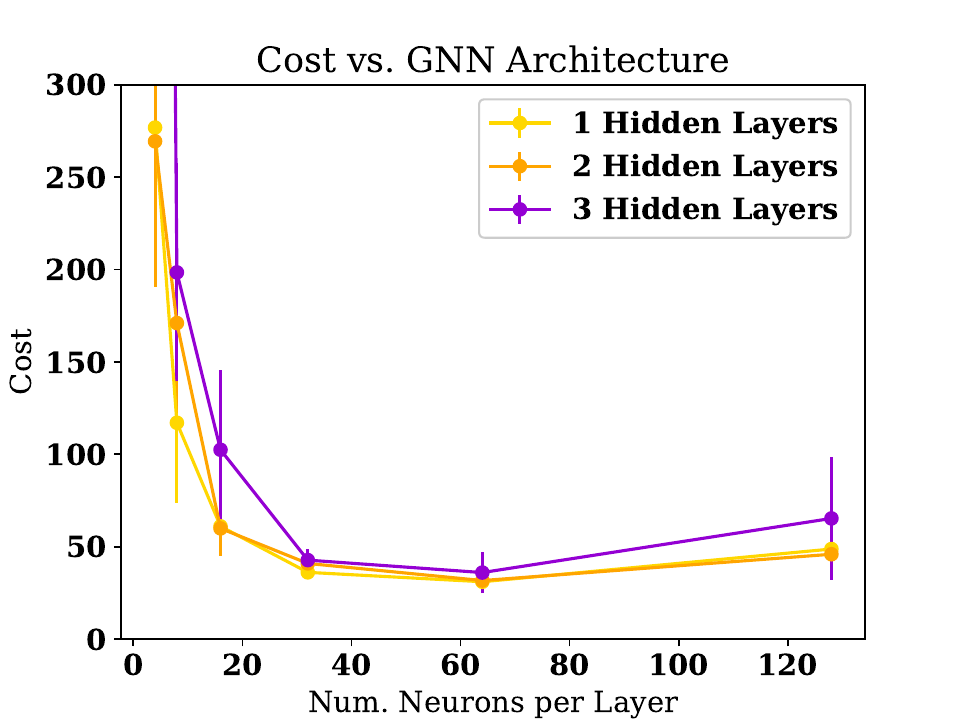} 
\subcaption{ } 
\label{fig:hidden_size}
\end{subfigure}%
\caption{ 
The flock's maximum initial velocities, communication radius, and the number of agents are key parameters affecting the control cost. The GNN architecture uses 2 hidden layers with 32 neurons each.
\vspace{-0.5cm}
}
\label{fig:flocking_benchmarking}
\end{figure}

Next, we study the performance of the GNN controller (for different values of $K$) compared to both the local controller $\bbu_{i}^{\dagger}$ and the global controller $\bbu_{i}^{\ast}$ [eq. \eqref{eq:controller_global}] under different flocking scenarios; namely, different initial velocities (Fig.~\ref{fig:vel}), communication radius (Fig.~\ref{fig:rad}), and number of agents (Fig.~\ref{fig:flocking_n}). Then, we also consider different aggregation GNN architecture hyperparameters (Fig.~\ref{fig:hidden_size}. We recall that the performance is measured by the cost metric defined in  \eqref{eq:flockcost}. In general, we observe that the GNN cost is bounded by the global controller as a best-case baseline, and the the local controller as the worst-case baseline, as expected, showing, in many cases, a marked improvement over the local controller $\bbu_{i}^{\dagger}$.


More specifically, in Fig. \ref{fig:vel}, we fix the communication radius $R = 1.0$ m for $N=100$ agents. The $K=1$ controller, which uses the same data as the local controller, performs an order of magnitude worse than $K=2$ to $K=4$, whose performance is comparable. At the highest initial velocity of $v_{init}=4.5$ m/s, the $K=4$ GNN performs slightly better than the rest.  Fig. \ref{fig:rad} shows that the flocking problem becomes more challenging as the communication radius decreases, for fixed maximum initial velocities, $v_{init} = 3.0$ m/s, and $N=100$ agents. The difference between the GNNs is most dramatic at $R=1.0 $ m, where the $K=3$ and $4$ controllers perform much better than all but the global controller. It is interesting to note that the cost of the local controller dips below the GNN controllers for large communication radii such as $R=4.0$ m. This may be due to the lack of truncation of the GNN features in  \eqref{eq:flockfeatures}. In Fig. \ref{fig:flocking_n}, we observe that the flocking cost per agent, computed by  \eqref{eq:flockcost}, decreases in value and variance as the flock size increases, with fixed $R = 1.0$ m and $v_{max}=3.0$ m/s. The GNN approach generalizes easily to 150 agents, with no penalties on larger flocks. In Fig. \ref{fig:hidden_size}, we observe that the GNN architecture resulting in the lowest control cost uses 1-2 hidden layers and 32-64 neurons, for fixed $v_{init} = 3.0$, $N=100$, $R = 1.0$ m. For smaller architectures, we observe under-fitting.

 
 \subsection{Transfer to Leader Following} \label{subsec:transfer}
 
Next, we investigate a new application of flocking in the presence of leader agents, with which the rest of the flock must align velocities. This application is extremely relevant for human-robot interaction for control of large swarms, and has not been previously explored by \cite{tanner2003stable}. 
In the previous section, the flock positions and velocities are initialized at random. This induces a symmetry in the configuration - the distribution of velocities in an agent's 1-hop neighborhood is the same as in its 2-hop or 3-hop neighborhood. Now, our goal is to investigate applications in which this symmetry is not present, to emphasize the capabilities of the aggregation GNN for $K=3$ and $K=4$. All models in this experiment were trained for $v_{init}=3.0$ m/s, $N=100$, and $R=1.0$ m. 
 
\begin{figure}[t]
\centering
  \setlength{\abovecaptionskip}{-2pt}
\begin{subfigure}[b]{.4\linewidth}
\includegraphics[width=\textwidth]{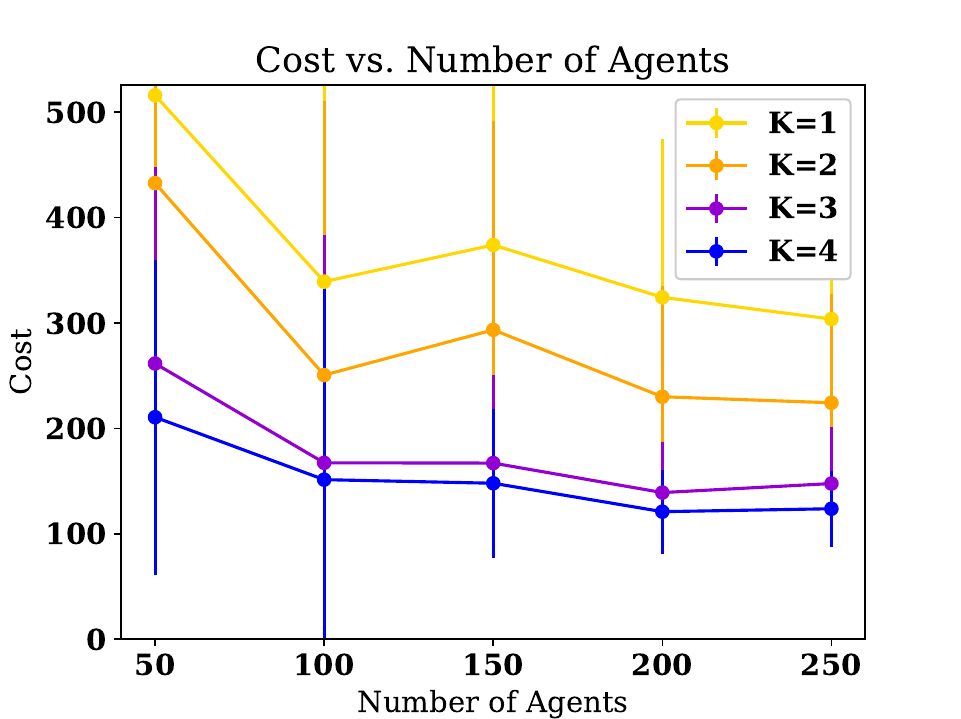} \subcaption{ 
Transfer to flocking with leaders.
}\label{fig:transfer_leader}
\end{subfigure}%
\begin{subfigure}[b]{.4\linewidth}
\includegraphics[width=\textwidth]{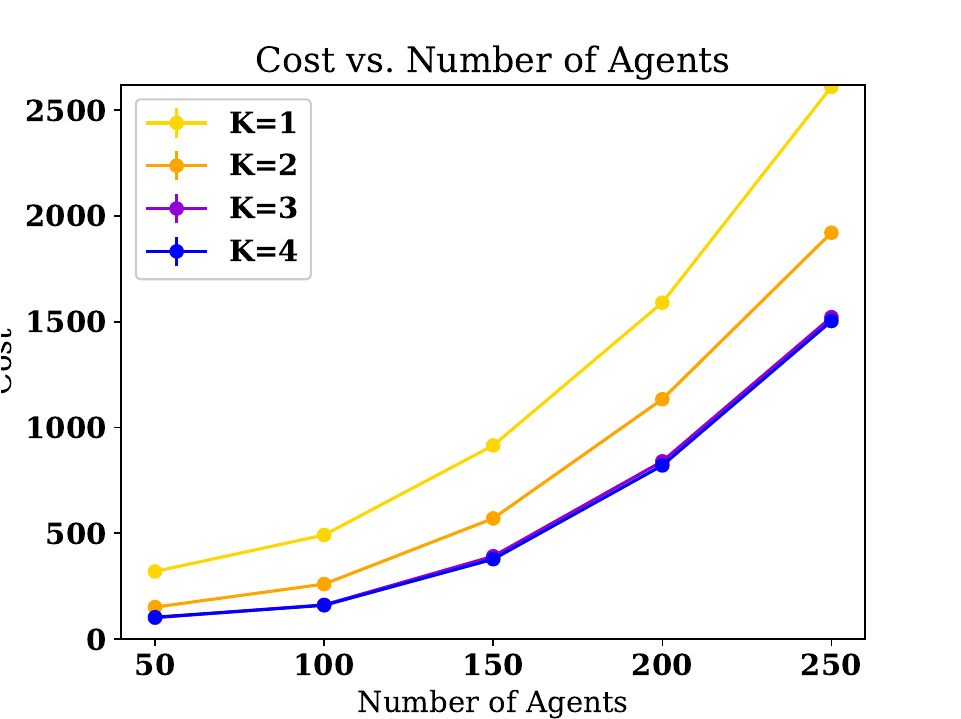}
\subcaption{ 
Transfer to grid formation.
}\label{fig:transfer_grid}
\end{subfigure}
\begin{subfigure}[b]{.4\linewidth}
\abovecaptionskip -2pt
\includegraphics[width=\textwidth]{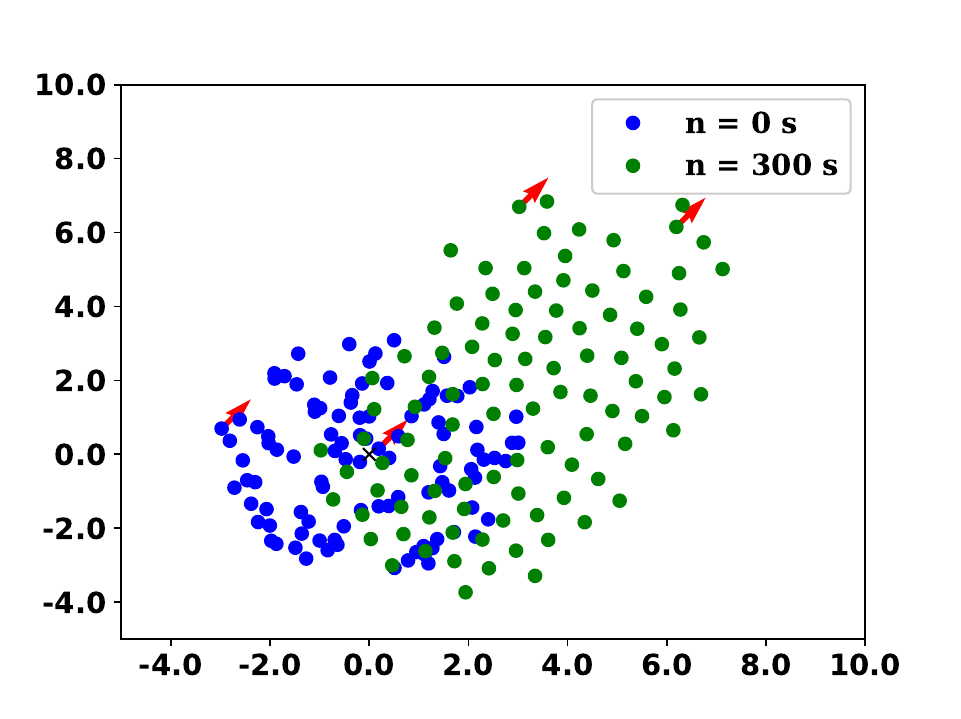}
\caption{Flocking with two leader agents}\label{fig:leader_traj_300}
\end{subfigure}
\begin{subfigure}[b]{.4\linewidth}
\abovecaptionskip -2pt
\includegraphics[width=\textwidth]{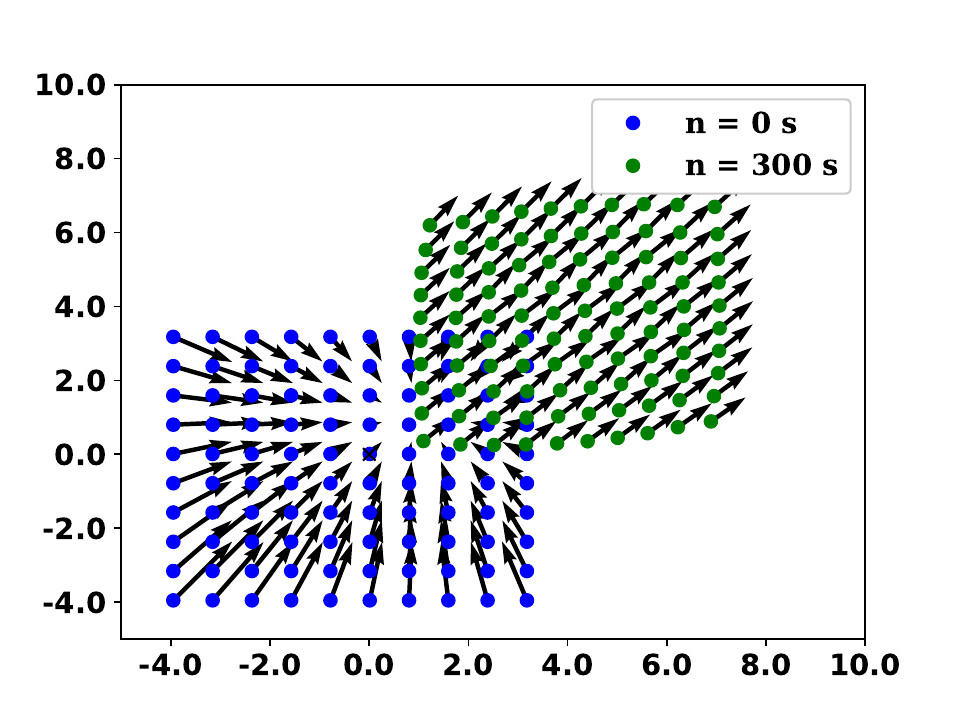}
\caption{Grid of agents with radial velocities}\label{fig:grid_traj300}
\end{subfigure}
\caption{ 
The trained GNN is transferred to new challenging scenarios with large numbers of agents.
\vspace{-0.5cm}
}
\label{fig:flocking_transfer}
\end{figure}

We first examine the transfer of the trained controllers to a system with two leader agents that have equal velocities that remain constant throughout an episode. That is, we train the architecture on examples of the symmetric network, but we test it on the leader system. Results are shown in Fig. \ref{fig:leader_traj_300}. We can observe that the $K=3$ and $K=4$ aggregation GNNs provide the best performance which is in agreement with the intuition that 2-hop and 3-hop aggregations allow the agents to respond more quickly to a leader agent who may be multiple hops away. 
In the second scenario, agents are initialized in a grid, with velocities radially inwards toward the centroid of the flock. The initial velocities of agents are initialized to be proportional to their distance from the center of the flock. Results are shown in Fig. \ref{fig:transfer_grid}. As the number of agents increases, we see the cost increases exponentially because the initial velocities increase, which is consistent with Fig. \ref{fig:vel}. This scenario is particularly prone to the scattering behavior of Fig. \ref{fig:dec_traj_300} due to the high probability of near-collisions, but the $K=3$ and $K=4$ controllers successfully align the agents' velocities and promote regular spacing among agents.


%% file: conclusion.tex
\section{Conclusion}

We have demonstrated the utility of aggregation graph neural networks as a tool for automatically learning distributed controllers for large teams of agents with coupled state dynamics and sparse communication links.
We envision the use of an Aggregation GNN-based controller in large-scale drone teams deployed into communication-limited environments to accomplish coverage, surveillance or mapping tasks.
In these settings, it is critical for agents to incorporate information from distant teammates in spite of local communication constraints; we show aggregations GNNs can be extended to accomplish this even with the time-varying agent states and time-varying communication networks typical of mobile robots.
In experiments, learning decentralized controllers for flocking and additional applications confirms the value of multi-hop information to performance and robustness to number of agents and communication radius.
In future work, enforcing state or input constraints could help avoid these failure modes.




%% file: appendix.tex
\section{Appendix}
\subsection{Aggregation GNN Evaluation}\label{sec_gnn_details}

%
\begin{figure*}[h]
    \centering
    \begin{subfigure}{0.200\textwidth}
        \centering
        \includegraphics[width=\textwidth]{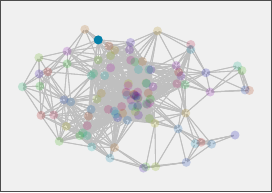}
        \caption{$\bby_{0n}$}
        \label{Agg_t0}
    \end{subfigure}
    \hfill
    \begin{subfigure}{0.200\textwidth}
        \centering
        \includegraphics[width=\textwidth]{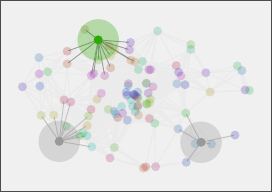}
        \caption{$\bby_{1n}$}
        \label{Agg_t1}
    \end{subfigure}
    \hfill
    \begin{subfigure}{0.200\textwidth}
        \centering
        \includegraphics[width=\textwidth]{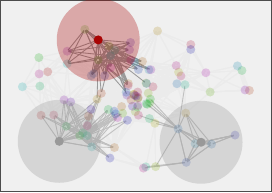}
        \caption{$\bby_{2n}$}
        \label{Agg_t2}
    \end{subfigure}
    \hfill
    \begin{subfigure}{0.200\textwidth}
        \centering
        \includegraphics[width=\textwidth]{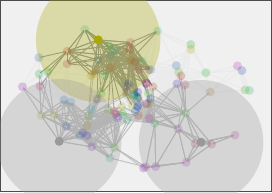}
        \caption{$\bby_{3n}$}
        \label{Agg_t3}
    \end{subfigure}
    \\ \vspace{0.67cm}
    \begin{subfigure}{1\textwidth}
        \centering
        \input{fig/atNode.tex}
        \caption{Processing at the $i$-th node}
        \label{atNode}
    \end{subfigure}
\caption{Aggregation graph neural networks perform successive local exchanges between each node and its neighbors. For each $k$-hop neighborhood (illustrated by the increasing disks), record $\bby_{kn}$  \eqref{eq:ydef} to build signal $\bbz_{n}$ which exhibits a regular structure \eqref{eq:z_def}. Figures \subref{Agg_t0} The value of the state \subref{Agg_t1} One-hop neighborhood \subref{Agg_t2} Two-hop neighborhood \subref{Agg_t3} Three-hop neighborhood. \subref{atNode} Once the regular time-structure signal $\bbz_{n}$ is obtained, we take each row $\bbz_{in}$, representing the information collected at node $i$, and we process it through a CNN to obtain the value of the decentralized controller $\bbu_{in}$ at node $i$ at time $n$ \eqref{eq:AggGNN}. }
	\label{fig_aggregation}
\end{figure*}

In this appendix, we present a detailed algorithm to compute the output of Aggregation GNNs, see Fig.~\ref{fig_aggregation}. Essentially, each node in the network exchanges information with its neighbors repeatedly, building sequence $\bbz_{in}$ \eqref{eq:z_def}. Since this sequence is a regular sequence (i.e. nearby entries in $\bbz_{in}$ correspond to the information aggregated from neighboring nodes), a traditional CNN can be directly applied, and an output computed. This CNN is applied locally at each node, with parameters shared across all nodes.

More specifically, Algorithm  \ref{algo_controller} summarizes the inference methodology for the aggregation GNN at a single node of the network. At time $n$, the agent receive aggregation sequences from its neighbors $z_{j(n-1)}$. Then, the agent pools this information from neighbors to form the current aggregation vector $z_{in}$ which is input to the learned controller $\pi(\bbz_{in}, \bbH)$ to compute the new action $\bbu_{in}$. Finally, the agent transmits its aggregation vector $\bbz_{in}$ to its current neighbors $\mathcal{N}_{in}$.

\begin{algorithm}[h]  %
\begin{algorithmic}[1]{\small
\For {n=0,1,\ldots,}
\State Receive aggregation sequences from $j\in\ccalN_{in}$
       [eq. \eqref{eq:z_def}]
       $$\bbz_{j(n-1)} = \Big[ \big[\bby_{0(n-1)}\big]_j \,;
                               \big[\bby_{1(n-1)}\big]_j \,;
                               \ldots\,; 
                               \big[\bby_{(K-1)(n-1)}\big]_j \Big]$$
\State Update aggregation sequence components 
       [eq. \eqref{eq:ydef} and \eqref{eq:localCommunication}]
       $$ \Big[\, \bby_{kn} \,\Big]_{i} \ =
           \Big[\,\bbS_{n} \bby_{k(n-1)}\,\Big]_{i} \ = 
                \sum_{j=1, j \in \ccalN_{in}} 
                     \Big[\,\bbS_{n}\,\Big]_{ij}\Big [\,\bby_{k(n-1)}\,\Big]_{j} \ $$
\State Observe system state $\bbx_{in}$
\State Update local aggregation sequence [cf. \eqref{eq:z_def}]
       $$\bbz_{in} = \Big[  \bbx_{in}^\Tr \,;
                             \big[\bby_{1n}\big]_i \,;
                             \ldots\,; 
                             \big[\bby_{(K-1)n}\big]_i \Big]$$
\State Compute next local action using the learned controller  $$\bbu_{in} = \pi\big(\bbz_{in}, \bbH\big)$$
\State Transmit local aggregation sequence $\bbz_{in}$ to neighbors $j\in\ccalN_{in}$
\EndFor}
\end{algorithmic}
\caption{Aggregation Graph Neural Network at Agent $i$.}\label{algo_controller}
\end{algorithm}

\subsection{High order dynamics} \label{subsec:airsim}

The ideal point masses provide a convenient benchmark for testing the effects of system parameters, but ultimately, we are most interested in flocking for robotic systems. Our goal in this section is compare several approaches to flocking in large teams of up to 50 quadrotors in simulation. Testing in the AirSim simulator allows us to test our controllers in the presence of higher order dynamics, slower control rates, and latency in observations \cite{airsim2017fsr}. 

The first challenge is that the AirSim simulator runs in real time, and produces an order of magnitude fewer training data points as compared to the point-mass simulation. To speed up training, we attempt to train on a simulated point mass system that was tuned to the parameters of AirSim. Since, we were not able to achieve an exact control frequency for the simulation, we instead observed that commands were issued at a variable time interval, with a mean of $0.12$ seconds, and a standard deviation of $0.018$ seconds.  Therefore, we adapted the point mass simulation to match the AirSim simulation by sampling $T_s \sim \mathcal{N}(0.12, 3\times10^{-4})$ and applying the linear model in \eqref{eq:discreteDynamic}. 
The inputs and outputs of the controllers were scaled by a factor of $l=6$ before evaluation of the GNN so that the optimal spacing dictated by the potential function \eqref{eq:potentialij} does not result in collisions. The scaling of the control actions and features was performed by:
$
\bbx_{n+1} = f(\bbx_n, l \cdot \pi(\bbx_n/l)),
$ where $f$ is the dynamics of the system, and $l=6$ is the scaling factor.

The second challenge was converting from the desired acceleration values produced by the controller to the desired roll and pitch commands. We follow the approach of \cite{mellinger2012trajectory}, which linearizes the dynamics of the robot about the hover point. In our case, the roll, $\phi$, and pitch, $\theta$, are proportional to the desired acceleration in the $x$ and $y$ dimensions. For a fixed yaw orientation, we used the approximation: $\phi = \bbu_2 / g$ and $\theta = - \bbu_1 / g$. The signs are inconsistent with those of \cite{mellinger2012trajectory} due to the AirSim convention that the $+Z$ dimension is down.
This approximation is only valid for small accelerations, so the acceleration inputs were clipped to a range of $[-3.0, 3.0]$ m/s$^2$.  Performance of the controllers was quantified using the variance of velocities in \eqref{eq:flockcost}. 
We initialize the flock uniformly spaced in a grid formation with a distance of about $4.8$ m between agents, and with x/y velocities sampled uniformly on the range $[-3.0, 3.0]$ m/s. The communication radius in AirSim was fixed at 9 m. 

\begin{figure}[h!]
\centering
\includegraphics[width=0.5\textwidth]{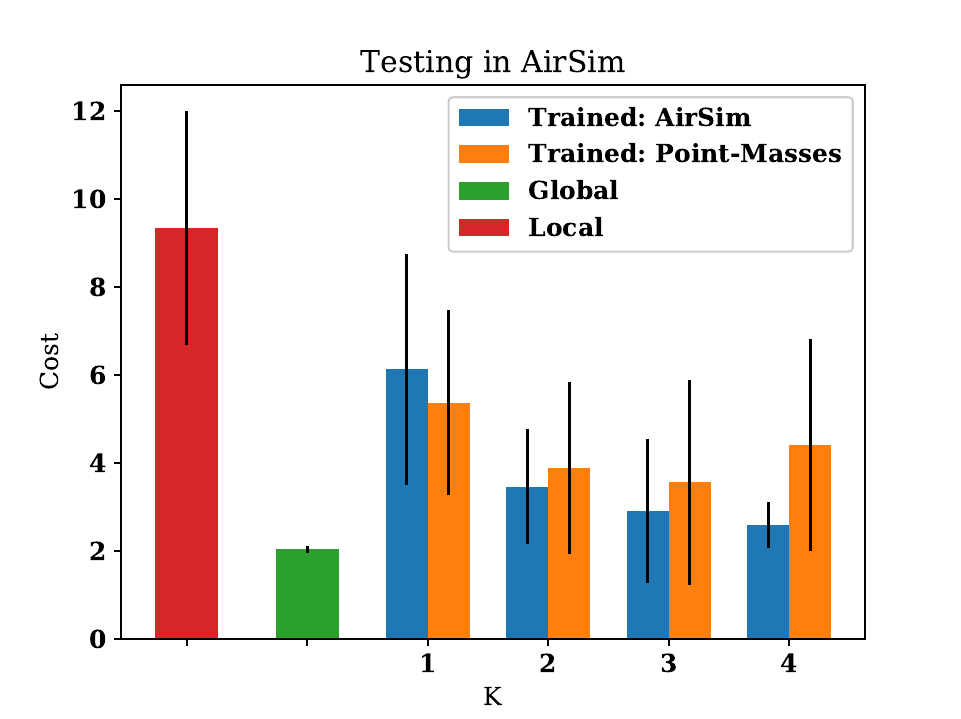} 
\caption{ 
Four models were trained in AirSim and four on the stochastic point masses, and then all models were tested in AirSim.
}
\label{fig:airsim_transfer}
\end{figure}

Fig. \ref{fig:airsim_transfer} compares the performance of controllers trained in simulation against those trained on the stochastic point mass model. The results of all eight GNN models are framed by the local and global controller as the upper and lower baselines. The GNN controller with $K=4$ trained in AirSim outperformed all other learned controllers. For $K=1$ and $K=2$, the controllers trained in simulation and AirSim had similar performance, but as additional aggregation operations are added, the benefit of training in AirSim becomes more obvious. We believe that the $K=4$ model trained on point masses has overfit to the ideal dynamics, so the performance degrades. 
 Both in simulation and in point-mass experiments, we observed a failure mode, in which a small group of agents is moving too quickly and escapes from the rest of the group. This small sub-flock typically exhibits flocking behavior among the several agents, but has no ability to re-join the flock, because it is permanently outside of the communication range of the rest of the agents. This drawback results from the lack of hard constraints on the connectivity in the system.

%% file: fig/atNode.tex

\def \thisplotscale {0.4}
\def \unit {\thisplotscale cm}

\tikzstyle {block} = [draw,
                      rectangle, 
                      fill = blue!10, 
                      minimum height = 2*\unit]

\tikzstyle {long block}  = [block,
                            minimum width = 4*\unit]

\tikzstyle {short block} = [block,
                            minimum width = 2*\unit]

\tikzstyle {circle block}   = [draw,
                            circle,
                            minimum width = 0.9*\unit]

\def\blockDistance{7}

{\footnotesize\begin{tikzpicture}[x = 1*\unit, y=1*\unit, font=\footnotesize]

\node at (0,0) (O) {};

%
\path(O) node (zn) {\includegraphics[width=0.2300\textwidth]{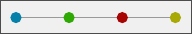}};
\path (zn.east) ++ (\blockDistance,0) node [long block] (CNN) {CNN};

%
\path [draw, -stealth] (zn.east) -- (CNN.west)  node  [midway, above] (z) {$\bbz_{in}$};
\path [draw, -stealth] (CNN.east) -++ (0.5*\blockDistance,0)  node  [midway,above] (outCNN) {$\bbu_{in}$};

\end{tikzpicture}}

%% file: main.bbl
\begin{thebibliography}{35}
\providecommand{\natexlab}[1]{#1}
\providecommand{\url}[1]{\texttt{#1}}
\expandafter\ifx\csname urlstyle\endcsname\relax
  \providecommand{\doi}[1]{doi: #1}\else
  \providecommand{\doi}{doi: \begingroup \urlstyle{rm}\Url}\fi

\bibitem[Sharma et~al.(2016)Sharma, Bennis, and Kumar]{sharma2016uav}
V.~Sharma, M.~Bennis, and R.~Kumar.
\newblock {UAV}-assisted heterogeneous networks for capacity enhancement.
\newblock \emph{IEEE Communications Letters}, 20\penalty0 (6):\penalty0
  1207--1210, 2016.

\bibitem[Thrun et~al.(2000)Thrun, Burgard, and Fox]{thrun2000real}
S.~Thrun, W.~Burgard, and D.~Fox.
\newblock A real-time algorithm for mobile robot mapping with applications to
  multi-robot and {3D} mapping.
\newblock In \emph{Robotics and Automation, 2000. Proceedings. ICRA 2000. IEEE
  International Conference on}, volume~1, pages 321--328. IEEE, 2000.

\bibitem[Thrun and Liu(2005)]{thrun2005multi}
S.~Thrun and Y.~Liu.
\newblock Multi-robot slam with sparse extended information filers.
\newblock In \emph{Robotics Research. The Eleventh International Symposium},
  pages 254--266. Springer, 2005.

\bibitem[Baxter et~al.(2007)Baxter, Burke, Garibaldi, and
  Norman]{baxter2007multi}
J.~L. Baxter, E.~Burke, J.~M. Garibaldi, and M.~Norman.
\newblock Multi-robot search and rescue: A potential field based approach.
\newblock In \emph{Autonomous robots and agents}, pages 9--16. Springer, 2007.

\bibitem[Jennings et~al.(1997)Jennings, Whelan, and
  Evans]{jennings1997cooperative}
J.~S. Jennings, G.~Whelan, and W.~F. Evans.
\newblock Cooperative search and rescue with a team of mobile robots.
\newblock In \emph{Advanced Robotics, 1997. ICAR'97. Proceedings., 8th
  International Conference on}, pages 193--200. IEEE, 1997.

\bibitem[Zhang and Hou(2005)]{zhang2005maintaining}
H.~Zhang and J.~C. Hou.
\newblock Maintaining sensing coverage and connectivity in large sensor
  networks.
\newblock \emph{Ad Hoc \& Sensor Wireless Networks}, 1\penalty0 (1-2):\penalty0
  89--124, 2005.

\bibitem[Witsenhausen(1968)]{witsenhausen1968counterexample}
H.~S. Witsenhausen.
\newblock A counterexample in stochastic optimum control.
\newblock \emph{SIAM Journal on Control}, 6\penalty0 (1):\penalty0 131--147,
  1968.

\bibitem[Pomerleau(1989)]{pomerleau1989alvinn}
D.~A. Pomerleau.
\newblock Alvinn: An autonomous land vehicle in a neural network.
\newblock In \emph{Advances in neural information processing systems}, pages
  305--313, 1989.

\bibitem[Giusti et~al.(2016)Giusti, Guzzi, Ciresan, He, Rodr{\'\i}guez,
  Fontana, Faessler, Forster, Schmidhuber, Di~Caro, et~al.]{giusti2016machine}
A.~Giusti, J.~Guzzi, D.~C. Ciresan, F.-L. He, J.~P. Rodr{\'\i}guez, F.~Fontana,
  M.~Faessler, C.~Forster, J.~Schmidhuber, G.~Di~Caro, et~al.
\newblock A machine learning approach to visual perception of forest trails for
  mobile robots.
\newblock \emph{IEEE Robotics and Automation Letters}, 1\penalty0 (2):\penalty0
  661--667, 2016.

\bibitem[Bruna et~al.(2014)Bruna, Zaremba, Szlam, and
  LeCun]{bruna14-deepspectralnetworks}
J.~Bruna, W.~Zaremba, A.~Szlam, and Y.~LeCun.
\newblock Spectral networks and locally connected networks on graphs.
\newblock In \emph{International Conference on Learning Representations
  ({{ICLR}})}, {Banff}, Apr. 2014.
\newblock URL \url{http://arxiv.org/abs/1213.6203}.

\bibitem[Defferrard et~al.(2016)Defferrard, Bresson, and
  Vandergheynst]{defferrard17-cnngraphs}
M.~Defferrard, X.~Bresson, and P.~Vandergheynst.
\newblock Convolutional neural networks on graphs with fast localized spectral
  filtering.
\newblock In \emph{Annu. Conf. Neural Inform. Process. Syst. 2016}, Barcelona,
  Spain, 5-10 Dec. 2016. NIPS Foundation.

\bibitem[Kipf and Welling(2017)]{kipf17-classifgcnn}
T.~N. Kipf and M.~Welling.
\newblock Semi-supervised classification with graph convolutional networks.
\newblock In \emph{5th Int. Conf. Learning Representations}, Toulon, France,
  24-26 Apr. 2017. Assoc. Comput. Linguistics.

\bibitem[Gama et~al.(2019)Gama, G.~Marques, Leus, and
  Ribeiro]{gama2018convolutional}
F.~Gama, A.~G.~Marques, G.~Leus, and A.~Ribeiro.
\newblock Convolutional neural network architectures for signals supported on
  graphs.
\newblock \emph{{IEEE} Trans. Signal Process.}, 67\penalty0 (4):\penalty0
  1034--1049, Feb. 2019.

\bibitem[Ruiz et~al.(2019)Ruiz, Gama, G.~Marques, and Ribeiro]{ruiz19nonlinear}
L.~Ruiz, F.~Gama, A.~G.~Marques, and A.~Ribeiro.
\newblock Median activation functions for graph neural networks.
\newblock In \emph{44th {IEEE} Int. Conf. Acoust., Speech and Signal Process.},
  Brighton, UK, 12-17 May 2019. IEEE.
\newblock URL \url{http://arxiv.org/abs/1810.12165}.

\bibitem[Tacchetti et~al.(2019)Tacchetti, Song, Mediano, Zambaldi, Kramar,
  Rabinowitz, Graepel, Botvinick, and Battaglia]{tacchetti2019relational}
A.~Tacchetti, H.~F. Song, P.~A.~M. Mediano, V.~Zambaldi, J.~Kramar, N.~C.
  Rabinowitz, T.~Graepel, M.~Botvinick, and P.~W. Battaglia.
\newblock Relational forward models for multi-agent learning.
\newblock In \emph{International Conference on Learning Representations}, {New
  Orleans}, May 2019.

\bibitem[Paulos et~al.(2019)Paulos, Chen, Shishika, and
  Vijay]{paulos2019decentralization}
J.~Paulos, S.~W. Chen, D.~Shishika, and K.~Vijay.
\newblock Decentralization of multiagent policies by learning what to
  communicate.
\newblock In \emph{2019 {{IEEE International Conference}} on {{Robotics}} and
  {{Automation}} ({{ICRA}})}, {Montreal}, May 2019.

\bibitem[H{\"u}ttenrauch et~al.(2019)H{\"u}ttenrauch, {\v S}o{\v s}i{\'c}, and
  Neumann]{huttenrauch2019deep}
M.~H{\"u}ttenrauch, A.~{\v S}o{\v s}i{\'c}, and G.~Neumann.
\newblock Deep {{Reinforcement Learning}} for {{Swarm Systems}}.
\newblock \emph{Journal of Machine Learning Research}, 20\penalty0
  (54):\penalty0 1--31, 2019.

\bibitem[Caro et~al.(2013)Caro, Giusti, Nagi, and Gambardella]{caro2013simple}
G.~A.~D. Caro, A.~Giusti, J.~Nagi, and L.~M. Gambardella.
\newblock A simple and efficient approach for cooperative incremental learning
  in robot swarms.
\newblock In \emph{2013 16th {{International Conference}} on {{Advanced
  Robotics}} ({{ICAR}})}, pages 1--8, Nov. 2013.
\newblock \doi{10.1109/ICAR.2013.6766596}.

\bibitem[Giusti et~al.(2012)Giusti, Nagi, Gambardella, and
  Caro]{giusti2012cooperative}
A.~Giusti, J.~Nagi, L.~Gambardella, and G.~A.~D. Caro.
\newblock Cooperative sensing and recognition by a swarm of mobile robots.
\newblock In \emph{2012 {{IEEE}}/{{RSJ International Conference}} on
  {{Intelligent Robots}} and {{Systems}}}, pages 551--558, Oct. 2012.
\newblock \doi{10.1109/IROS.2012.6385982}.

\bibitem[Otte(2018)]{otte2018emergent}
M.~Otte.
\newblock An emergent group mind across a swarm of robots: {{Collective}}
  cognition and distributed sensing via a shared wireless neural network.
\newblock \emph{The International Journal of Robotics Research}, 37\penalty0
  (9):\penalty0 1017--1061, Aug. 2018.
\newblock \doi{10.1177/0278364918779704}.

\bibitem[Tanner(2004)]{tanner2004flocking}
H.~G. Tanner.
\newblock Flocking with obstacle avoidance in switching networks of
  interconnected vehicles.
\newblock In \emph{IEEE International Conference on Robotics and Automation},
  volume~3, pages 3006--3011. Citeseer, 2004.

\bibitem[Reynolds(1987)]{reynolds1987flocks}
C.~W. Reynolds.
\newblock Flocks, {{Herds}} and {{Schools}}: {{A Distributed Behavioral
  Model}}.
\newblock In \emph{Proceedings of the 14th {{Annual Conference}} on {{Computer
  Graphics}} and {{Interactive Techniques}}}, {{SIGGRAPH}} '87, pages 25--34,
  {New York, NY, USA}, 1987. {ACM}.
\newblock \doi{10.1145/37401.37406}.

\bibitem[Jadbabaie et~al.(2003)Jadbabaie, Lin, and
  Morse]{jadbabaie2003coordination}
A.~Jadbabaie, J.~Lin, and A.~S. Morse.
\newblock Coordination of groups of mobile autonomous agents using nearest
  neighbor rules.
\newblock \emph{IEEE Transactions on automatic control}, 48\penalty0
  (6):\penalty0 988--1001, 2003.

\bibitem[Tanner et~al.(2003)Tanner, Jadbabaie, and Pappas]{tanner2003stable}
H.~G. Tanner, A.~Jadbabaie, and G.~J. Pappas.
\newblock Stable flocking of mobile agents part ii: dynamic topology.
\newblock In \emph{Decision and Control, 2003. Proceedings. 42nd IEEE
  Conference on}, volume~2, pages 2016--2021. IEEE, 2003.

\bibitem[Zhou et~al.(1996)Zhou, Doyle, Glover, et~al.]{zhou1996robust}
K.~Zhou, J.~C. Doyle, K.~Glover, et~al.
\newblock \emph{Robust and optimal control}, volume~40.
\newblock Prentice hall New Jersey, 1996.

\bibitem[Bemporad et~al.(2002)Bemporad, Morari, Dua, and
  Pistikopoulos]{bemporad2002explicit}
A.~Bemporad, M.~Morari, V.~Dua, and E.~N. Pistikopoulos.
\newblock The explicit linear quadratic regulator for constrained systems.
\newblock \emph{Automatica}, 38\penalty0 (1):\penalty0 3--20, 2002.

\bibitem[Eksin et~al.(2013)Eksin, Molavi, Ribeiro, and
  Jadbabaie]{eksin2013bayesian}
C.~Eksin, P.~Molavi, A.~Ribeiro, and A.~Jadbabaie.
\newblock Bayesian quadratic network game filters.
\newblock In \emph{Acoustics, Speech and Signal Processing (ICASSP), 2013 IEEE
  International Conference on}, pages 4589--4593. IEEE, 2013.

\bibitem[Sandryhaila and Moura(2014)]{sandryhaila14-mag}
A.~Sandryhaila and J.~M.~F. Moura.
\newblock Big data analysis with signal processing on graphs.
\newblock \emph{{IEEE} Signal Process. Mag.}, 31\penalty0 (5):\penalty0 80--90,
  Sep. 2014.

\bibitem[Chen et~al.(2015)Chen, Varma, Sandryhaila, and Kova{\v{c}}evi{\'
  c}]{chen15-selection}
S.~Chen, R.~Varma, A.~Sandryhaila, and J.~Kova{\v{c}}evi{\' c}.
\newblock Discrete signal processing on graphs: Sampling theory.
\newblock \emph{{IEEE} Trans. Signal Process.}, 63\penalty0 (24):\penalty0
  6510--6523, Dec. 2015.

\bibitem[G.~Marques et~al.(2016)G.~Marques, Segarra, Leus, and
  Ribeiro]{marques16-aggregation}
A.~G.~Marques, S.~Segarra, G.~Leus, and A.~Ribeiro.
\newblock Sampling of graph signals with successive local aggregations.
\newblock \emph{{IEEE} Trans. Signal Process.}, 64\penalty0 (7):\penalty0
  1832--1843, Apr. 2016.

\bibitem[Gama et~al.(2019)Gama, Bruna, and Ribeiro]{gama19-stability}
F.~Gama, J.~Bruna, and A.~Ribeiro.
\newblock Stability properties of graph neural networks.
\newblock \emph{arXiv:1905.04497v2 [cs.LG]}, 4 Sep. 2019.
\newblock URL \url{http://arxiv.org/abs/1905.04497}.

\bibitem[Xiao et~al.(2007)Xiao, Boyd, and Kim]{xiao2007distributed}
L.~Xiao, S.~Boyd, and S.-J. Kim.
\newblock Distributed average consensus with least-mean-square deviation.
\newblock \emph{Journal of parallel and distributed computing}, 67\penalty0
  (1):\penalty0 33--46, 2007.

\bibitem[Ross et~al.(2011)Ross, Gordon, and Bagnell]{ross2011reduction}
S.~Ross, G.~Gordon, and D.~Bagnell.
\newblock A reduction of imitation learning and structured prediction to
  no-regret online learning.
\newblock In \emph{Proceedings of the fourteenth international conference on
  artificial intelligence and statistics}, pages 627--635, 2011.

\bibitem[Shah et~al.(2017)Shah, Dey, Lovett, and Kapoor]{airsim2017fsr}
S.~Shah, D.~Dey, C.~Lovett, and A.~Kapoor.
\newblock Airsim: High-fidelity visual and physical simulation for autonomous
  vehicles.
\newblock In \emph{Field and Service Robotics}, 2017.

\bibitem[Mellinger(2012)]{mellinger2012trajectory}
D.~Mellinger.
\newblock \emph{Trajectory Generation and Control for Quadrotors}.
\newblock Ph.{{D}}., University of Pennsylvania, {Philadelphia}, 2012.

\end{thebibliography}
